\documentclass[a4paper, fleqn, times, review, 10pt]{cas-dc}

\usepackage[lined,boxed,commentsnumbered]{algorithm2e}
\usepackage{hyperref}
\usepackage{placeins}
\hypersetup{hidelinks,
colorlinks=true,
allcolors=black,
pdfstartview=Fit,
breaklinks=true}
\usepackage {algorithmic}
\usepackage{ulem}
\usepackage[authoryear,longnamesfirst]{natbib}
\usepackage{amssymb}
\usepackage{times}
\usepackage{soul}
\usepackage{url}
\usepackage{graphicx}
\usepackage{amsmath}
\usepackage{amsthm}
\usepackage{booktabs}
\usepackage{algorithmic}
\usepackage{multirow}
\def\tsc#1{\csdef{#1}{\textsc{\lowercase{#1}}\xspace}}
\tsc{WGM}
\tsc{QE}
\tsc{EP}
\tsc{PMS}
\tsc{BEC}
\tsc{DE}
\begin{document}
\let\WriteBookmarks\relax
\def\floatpagepagefraction{1}
\def\textpagefraction{.001}

\shorttitle{Temporal and Spatial Feature Fusion Framework for DMER}

\shortauthors{Feng Liu et al.}

\title [mode = title]{Temporal and Spatial Feature Fusion Framework for Dynamic Micro Expression Recognition}                      

% \tnotetext[1]{This document is the results of the research project funded by East China Normal University 2023 Humanities and Social Sciences Youth Pre-Research Project (2023ECNU-YYJO26) and Beijing Key Laboratory of Behavior and Mental Health, Peking University.}

%\tnotetext[2]{The second title footnote which is a longer text matter to fill through the whole text width and overflow into another line in the footnotes area of the first page.}


\author[1]{Feng Liu}[style=chinese, auid=000, bioid=1, orcid=0000-0002-5289-5761]
\cortext[cor1]{Corresponding author}
\cormark[1]
\ead{lsttoy@163.com/liu.feng@sjtu.edu.cn}

\author[2]{Bingyu Nan}[style=chinese, auid=000, bioid=1, orcid=0009-0001-7235-2617]    

\author[2]{Xuezhong Qian}[style=chinese, auid=000, bioid=2, orcid=0009-0001-7235-2617]    

\author[1]{Xiaolan Fu}[style=chinese, auid=000, bioid=1]
\cormark[1]
\ead{fuxl@psych.ac.cn}

\affiliation[1]{organization={School of Psychology,Shanghai Jiao Tong University},
    addressline={No.1954 Huashan Road, Xuhui District, Shanghai, China}, 
    city={Shanghai},
    citysep={}, % Uncomment if no comma needed between city and postcode
    postcode={200030}, 
    state={},
    country={China}}
    
\affiliation[2]{organization={Jiangnan University},
    addressline={No.1800 Lihu Avenue, Binhu District}, 
    city={wuxi},
    citysep={}, % Uncomment if no comma needed between city and postcode
    postcode={214122}, 
    state={},
    country={China}}

\begin{abstract}
When emotions are repressed, an individual's true feelings may be revealed through micro-expressions. Consequently, micro-expressions are regarded as a genuine source of insight into an individual's authentic emotions. However, the transient and highly localised nature of micro-expressions poses a significant challenge to their accurate recognition, with the accuracy rate of micro-expression recognition being as low as 50\%, even for professionals. In order to address these challenges, it is necessary to explore the field of dynamic micro expression recognition (DMER) using multimodal fusion techniques, with special attention to the diverse fusion of temporal and spatial modal features. In this paper, we propose a novel Temporal and Spatial feature Fusion framework for DMER (TSFmicro). This framework integrates a Retention Network (RetNet) and a transformer-based DMER network, with the objective of efficient micro-expression recognition through the capture and fusion of temporal and spatial relations. Meanwhile, we propose a novel parallel time-space fusion method from the perspective of modal fusion, which fuses spatio-temporal information in high-dimensional feature space, resulting in complementary “where-how” relationships at the semantic level and providing richer semantic information for the model. The experimental results demonstrate the superior performance of the TSFmicro method in comparison to other contemporary state-of-the-art methods. This is evidenced by its effectiveness on three well-recognised micro-expression datasets.

% Our code is accessible on Github. Our code could access at Github(\href{https://github.com/Cross-Innovation-Lab/TSFmicro}{https://github.com/Cross-Innovation-Lab/TSFmicro}).

\end{abstract}

% \includegraphics{figs/grabs.pdf}

\begin{highlights}

\item Proposed a novel dual-stream framework called TSFmicro for dynamic micro-expression recognition that effectively fuses time-dependent and spatial features.
\item Identified a parallel temporal and spatial fusion method that captures complementary “where-how” relationships in high-dimensional feature space to provide richer semantic information for models.
\item The feature contribution relationship between temporal and spatial modalities in the field of micro-expression recognition is revisited, and a new paradigm for pattern recognition research with multimodal fusion and spatio-temporal feature computation is further proposed.
\item The application of multimodal fusion techniques in dynamic micro-expression pattern recognition is explored, focusing on the diverse fusion of temporal and spatial modal features and their effects and contributions in different combinations of patterns.

\end{highlights}

% Keywords
% Each keyword is separated by \sep

\maketitle

\section{Introduction}
When emotions are repressed, an individual's authentic emotions will manifest through micro-expressions, which are therefore considered an important source of insight into an individual's authentic emotions\cite{CVPR2022zeng_fer,nn_wang2025,nn_gao2025,TMM2023ben}. This phenomenon is not governed by conscious thought, cannot be disguised or camouflaged, and can serve as an authentic reflection of an individual's inner emotional state. This feature enables micro-expressions to convey a person's true psychological state with a high degree of accuracy, thus possessing both research value and practical significance in clinical and national security domains\cite{pr_he2017,li2019blended,nn_zhao2022,CVPR2024Paskaleva_au,nn_tang2024}.

In the field of psychology, facial expressions are typically categorised into two primary types: macro-expressions and micro-expressions. Macro-expressions are distinguished by their high intensity and protracted duration, which typically extends between two and three seconds\cite{CVPR2023wang_DFER,MM2024li_DFER}. Their observation is often uncomplicated. In contrast, micro-expressions are distinguished by their low intensity, localised nature and brief duration (1/25 to 1/2 second)\cite{pr_ling2022,TIPli}. Furthermore, micro-expressions are spontaneous, unconscious facial activities that reveal people's authentic emotions\cite{MM2021yang,MM2024ge_au}. Nevertheless, accurate recognition of micro-expressions by the naked eye remains a significant challenge. Firstly, micro-expressions are characterised by their high temporal dynamics, which poses a significant challenge to the accurate recognition of micro-expressions using single-frame images\cite{aaai2023cmnet}. Secondly, micro-expressions are distinguished by their low motion intensity and their manifestation in specific localised areas\cite{TOMM2023gong}. This poses a challenge for accurate manual recognition of micro-expressions. Current research in this field demonstrates that the accuracy of micro-expression recognition is only 50\%\cite{2024review}, even for professionals. Consequently, there is a necessity to develop intelligent analysis of micro-expressions in order to more accurately capture the authentic emotions of individuals.The total data volume of the 12 published spontaneous micro-expressions datasets (CASME\cite{casme}, \\CASME \uppercase\expandafter{\romannumeral2}\cite{casme2}, CAS(ME)$^2$\cite{cas(me)2}, CAS(ME)$^3$\cite{casme3}, SMIC\cite{smic}, SMIC-E\cite{smice}, 4DME\cite{4dme}, SAMM\cite{samm}, SAMM-LV\cite{sammlv}, \\MMEW\cite{mmew}, DFME\cite{dfme} and MEVIEW\cite{MEVIEW}) is about 10,000, which is a typical classification and recognition task compared to other small-sample problem compared to other classification recognition tasks. This factor also has a significant impact on the development of deep learning in the field of micro-representation.

\begin{figure*}[!h]
\centering
\includegraphics[width=1\textwidth]{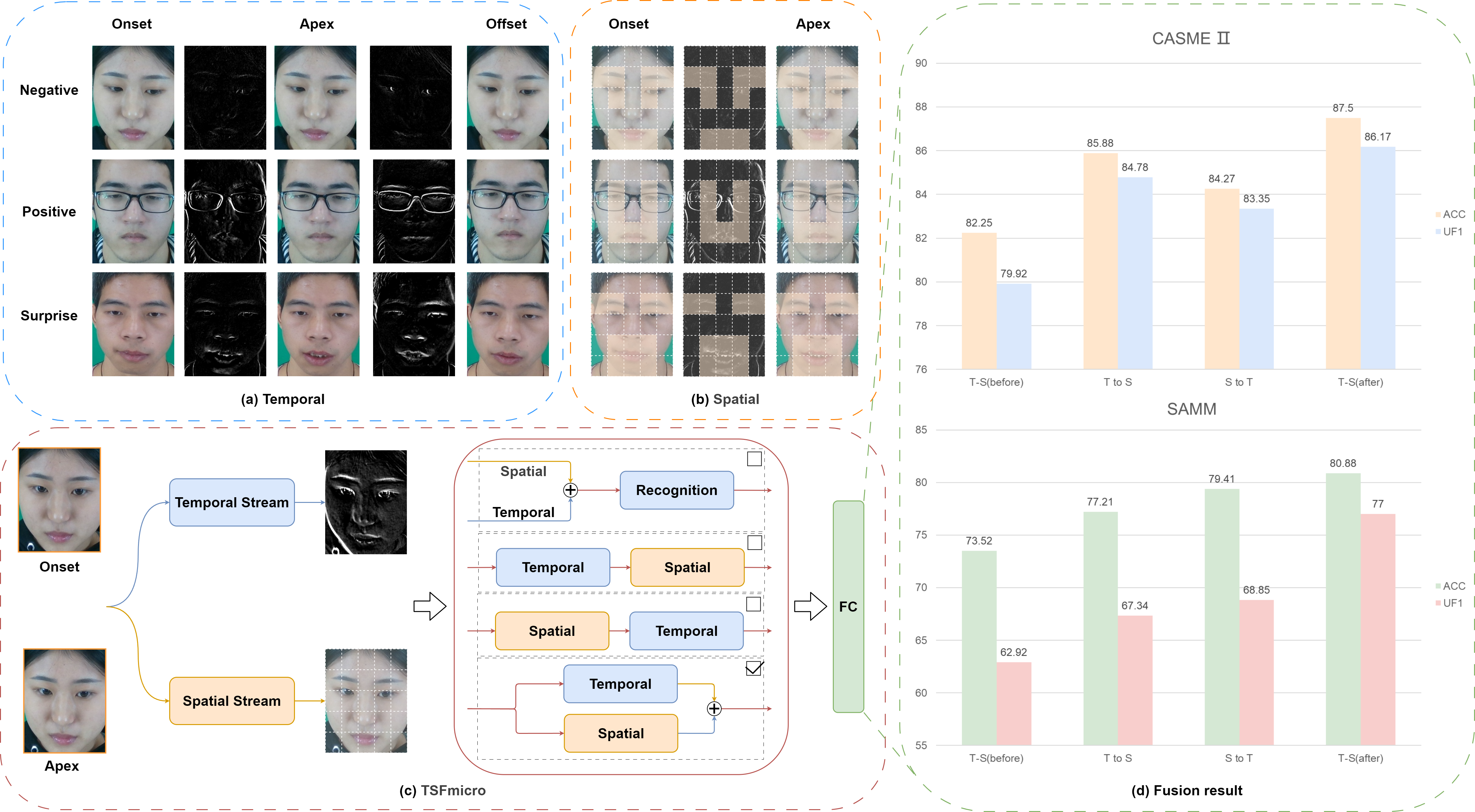}
\caption{This paper attempts to consider the potential impact of temporal fusion on micro-expression recognition performance from the perspective of modal fusion. (a) Temporal information between frames is extracted as a temporal feature using difference frames. (b) Position embedding is utilized to learn the positional information associated with the occurrence of an action in order to map it to temporal features. (c) Extract spatio-temporal information through temporal and spatial streams and experiment with different spatio-temporal fusion approaches. (d) Performance of TSFmicro with different spatio-temporal modal fusion approaches.}
\label{fig1}
\end{figure*}

Notwithstanding the attendant challenges, many different approaches have been tried over time to meet these challenges\cite{TMM2023chen,pr_gan2025,pr_zhang2022,pr_2014Majumder}. Initial MER studies concentrated on analysing complete time series, which resulted in data analysis that was both time-consuming and redundant. Subsequent researchers, however, analysed the process of micro-expression recognition from a psychological perspective. It was determined that the intensity of a micro-expression at its peak conveys information that reflects the current emotional state\cite{ekman1993}. Subsequent studies have experimentally verified this claim and proposed the use of Apex frames as the basis for the MER task, rather than complete micro-expression sequences\cite{liong2017}. Despite the notable success of micro-expression recognition methods based on single-frame images in achieving high accuracy, there is still scope for enhancement in terms of model accuracy when compared to other classification recognition tasks. In addition, micro-expressions are typically confined to specific localised areas of the face and tend to be brief in duration. For instance, the micro-expression of anger can be manifested through action units, i.e., the combination of AU4, AU5, AU7 and AU23\cite{CVPR2022chang_au}. This feature suggests that the model needs to understand the information about the specific location where the action occurs, so as to understand its temporal-spatio dynamic relationship more effectively. Furthermore, micro-expression sequences contain action change information, and this inter-frame action information can effectively facilitate the model's understanding of the temporal relationship.

% \begin{figure}[!htbp]
% \centering
% \includegraphics[width=0.95\textwidth]{figs/S.png}
% \caption{XXXX}
% \label{fig2}
% \end{figure}

To solve the above problem, we propose a novel dynamic temporal and spatial feature fusion framework for DMER, namely TSFmicro. it includes a temporal sub-branch and a spatial sub-branch. In the temporal branch, the action transformation information is captured by calculating the difference between Onset frames and Apex frames. Meanwhile, the spatial sub-branch is used to extract the location information of the action occurrence. Finally, we propose a novel parallel spatio-temporal fusion method, which fuses spatio-temporal information in high-dimensional feature space to form complementary “where-how” relationships at the semantic level, providing richer semantic information for the model.
As mentioned above, the core contributions of this paper can be listed as follows.

\begin{itemize}
 
\item Explores the field of dynamic micro expression pattern recognition using multimodal fusion techniques is explored, with particular attention paid to the diverse fusion of temporal and spatial modal features to clarify the effects and contributions of different modal features in various combined patterns.

\item A multimodal parallel fusion approach has been identified that can utilise spatio-temporal information in the fused high-dimensional feature space to form complementary "where-how" relationships at the semantic level. This approach provides richer semantic information for the model, and, in the spatial sub-branches, positional embedding is utilised to facilitate the acquisition of positional information related to the occurrence of the action while avoiding the acquisition of identity information unrelated to location information by reducing the number of network layers. The eventual fusion of temporal and spatial bimodal features is realised simultaneously in order to improve the dynamic micro-expression recognition effect.

\item it revisits the feature contribution relationship between temporal and spatial modalities in the field of micro-expression recognition, and further propose a new paradigm for pattern recognition research regarding the multimodal fusion and computation of spatio-temporal features.
 
\item A novel temporal-spatio feature fusion framework is proposed, which captures the dynamic temporal information by calculating the difference frames and maps them to the positional information extracted from the spatial sub-branches in order to achieve efficient integration of temporal-spatio information and experiments are conducted on three mainstream datasets(CASME \uppercase\expandafter{\romannumeral2}, SAMM and {CAS(ME)$^3$)} with results which exceed the State Of The Art(SOTA).

\end{itemize}

The subsequent sections of the paper are as follows: section 2 describes the related work; in section 3, we detail our proposed temporal-spatio fusion framework; in section 4, we describe the experimental procedure on the CASME \uppercase\expandafter{\romannumeral2}, CAS(ME)$^3$ and SAMM datasets; and in section 5, conclusions are drawn.

\section{Related Work}
In this section, we review previous work that is most related to this paper, such as existing MER approaches and frame selection methods.

\subsection{Micro-expression recognition in spatial perspective}

In the field of micro-expression recognition, spatial-based approaches focus on static features in micro-expressions to recognize emotions by analyzing the motion regions of facial muscles. Early studies relied heavily on manually extracted features, for example, one study proposed to perform the micro-expression recognition task by combining Local Binary Patterns in Three Orthogonal Planes (LBP-TOP) with a conventional classifier\cite{li2013lbp}. Based on this, other researchers further extended the original technique with the aim of reducing the redundancy of the LBP-TOP operator to improve the recognition accuracy\cite{wang2015}. In addition, a micro expression recognition method based on tensor-independent color space (TICS) has also been proposed, which significantly improves the accuracy of micro-expression recognition by combining LBP-TOP with color space\cite{wang2015color}. Although these methods show some effectiveness, they are limited by feature complexity and sensitivity to noise. In recent years, with the development of deep learning, convolutional neural networks have been widely used for micro-expression recognition. For example, there have been studies on recognizing micro-expressions through effective feature encoding and 2D convolutional neural networks\cite{Gupta2023}. There are also studies that implement objective category-based micro-expression recognition using region-based heuristic relational inference networks\cite{mao2022}. Although these approaches have made progress in recognizing static micro-expressions, they tend to ignore the epochal dynamics of micro-expressions and lack the ability to model temporal dynamic information.

\subsection{Micro-expression recognition in temporal perspective}

Unlike spatial-based approaches, time-based approaches emphasize the temporal dynamic characteristics of micro-expressions and identify emotions by analyzing changes in micro-expressions over time series. Earlier time-based methods mainly relied on optical flow features to capture the temporal dynamic features in micro-expressions by calculating the motion information between consecutive frames. For example, one study proposed a micro-expression recognition method based on the main-direction averaged optical flow (MDMO) feature, which achieves effective recognition of micro-expressions by counting the optical flow information in regions of interest (RoIs)\cite{liu2016flow}. In addition, a two-way weighted optical flow feature extractor has also been proposed to improve the performance of micro-expression recognition by selecting three key regions of interest (RoIs) associated with the micro-expression information and using optical flow strain and block-based LBP-TOP feature descriptors\cite{Liong2018}. Although these methods improve the micro-expression recognition performance, they are limited by noise and illumination variations and tend to exhibit low robustness. With the development of deep learning techniques, some studies have designed a shallow three-stream network architecture, which extracts three features, namely, optical strain, horizontal optical flow and vertical optical flow, by processing the start and peak frames of the input samples and combining them with optical flow features for classification\cite{liong2016ststnet}. In addition, other studies have further introduced attentional mechanisms combined with LSTM networks to capture micro-expression temporal dynamic features of facial sequences\cite{gan2023}. Despite the advantages of time-based methods in capturing the dynamic features of micro-expressions, micro-expressions are inherently localized in nature, and simply modeling the temporal sequences may overlook critical spatial features.

\subsection{Temporal-Spatio Fusion}

In the field of micro-expression recognition (MER), the development of frameworks that incorporate temporal and spatial dimensions has become a mainstream trend. In recent years, researchers have proposed a variety of innovative approaches. For example, some studies have introduced a variety of optical flow-derived components and OFF-ApexNet structures with a view to more accurately represent subtle facial motion changes\cite{liong2020gan}. Another study proposed a spatio-temporal LBP-TOP descriptor based on multi-scale active patch fusion that takes into account the unique contributions of different facial regions in micro-expression recognition\cite{sun2020lbptop}. In addition, there have been studies that have utilized self-higher order statistics of spatial and channel features to detect action units in micro-expression sequences\cite{li2021channel}. Based on this, one study proposed a dual-scale spatio-temporal feature learning method for micro-expression recognition using LSTM\cite{kim2016cnnlstm}. Other researchers have proposed a spatio-temporal recurrent convolutional network based on two different temporal connections to model the spatio-temporal deformation of micro-expression sequences\cite{xia2020}. In addition, studies have also proposed a spatio-temporal neural architecture search algorithm that implements end-to-end spatio-temporal feature learning through parallelogram structure search space and 3D convolution\cite{verma2022}. Despite the success of these approaches in achieving significant results in micro-expression recognition, most of them focus on the capture and modeling of single modal features, and lack of thinking about the potential impact of spatio-temporal fusion on micro-expression recognition performance from the perspective of modal fusion. In addition, a key issue that has not been addressed is how to prevent spatial sub-branches from extracting identity information that is not related to micro-expression motion.

Unlike them, this paper performs dynamic micro-expression pattern recognition by utilizing multimodal fusion techniques and pays special attention to the multivariate fusion of spatio-temporal modal features in order to elucidate the roles and contributions of different modal features in various combined patterns. Meanwhile, in the spatial sub-branch, fewer network layers are utilized to avoid acquiring identity information that is not related to location information.

\begin{figure*}[!t]
\centering
\includegraphics[width=0.95\textwidth]{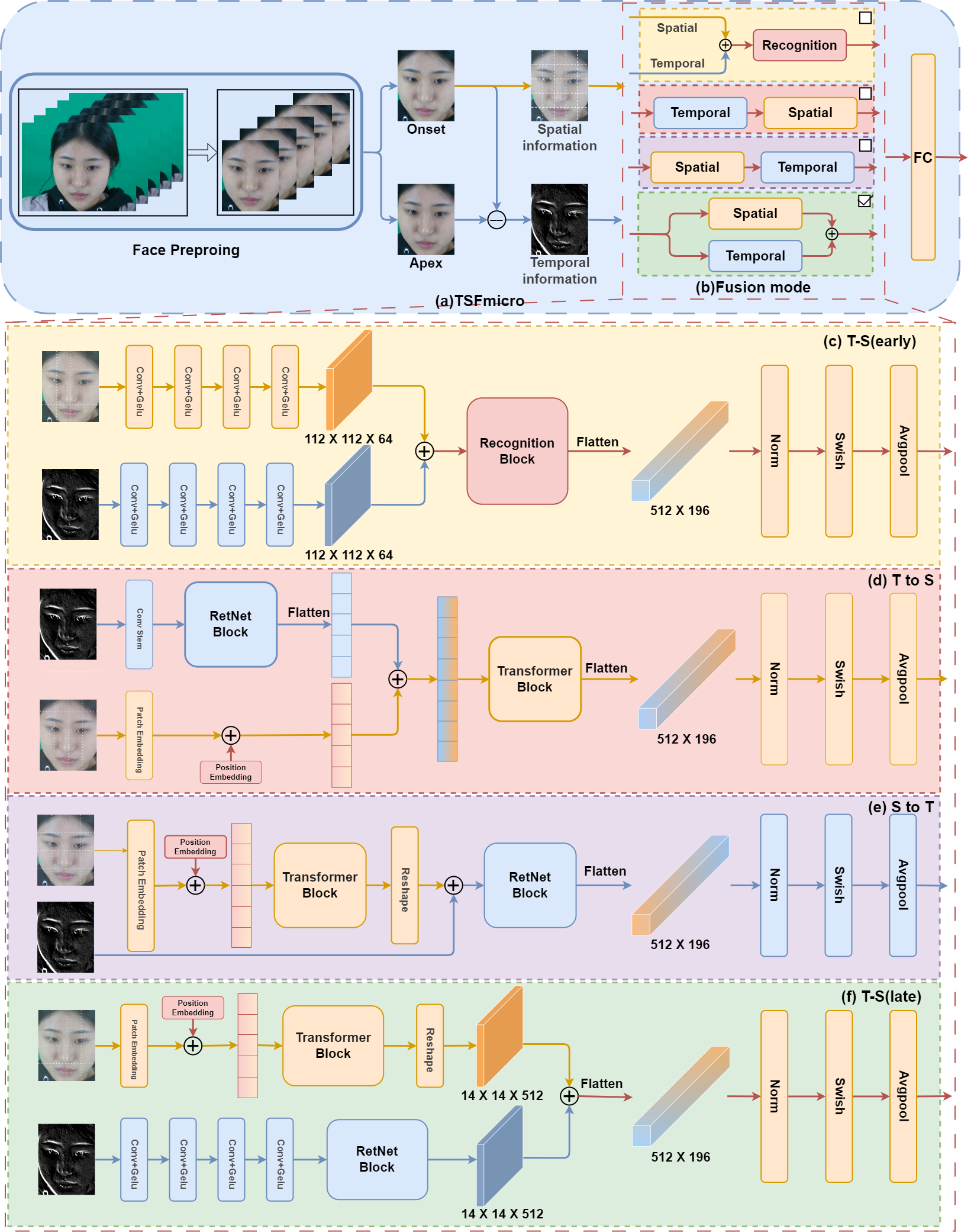}
\caption{An overview of the proposed TSFmicro architecture is presented below. (a)The process of TSFmicro is outlined as follows: firstly, the face is cropped; secondly, the difference frames between Apex and Onset frames are used as the temporal information and Onset frames are used as the spatial information; thirdly, the spatio-temporal sub-branch captures and fuses the spatio-temporal information; and finally, the data is categorized. (b) Fusion module. (c) The structure of the T to S (early) fusion approach. (d) The structure of the T to S fusion approach. (e) Structural delineation of the S-to-T fusion approach. (f) The structure of the T-S (late) fusion approach.}
\label{fig2}
\end{figure*}

\section{Methodology}

In this section, a novel framework for the integration of temporal and spatial features, designated TSFmicro, is proposed.As illustrated in Figure \ref{fig2}, the framework comprises temporal and spatial sub-branches. For the temporal sub-branch, the retention mechanism in Retnet is utilised to introduce temporal decay into the sequence processing in order to extract the action changes in the micro-expression sequences. For the spatial sub-branch, the initial procedure involves slicing the Onset frames and introducing positional information for each slice using the property of Position Embedding. Concurrently, a shallow network is built to extract the spatial features of micro-expressions, there by ensuring that identity information unrelated to spatial features is not extracted. Finally, we explore the effects of three different temporal-spatio feature fusion methods on the micro-expression recognition performance.

\subsection{Temporal sub-branch}

Onset frames are defined as the initial frames marking the onset of the micro-expression movement, while Apex frames are represented by the peak frames at which the facial micro-expression reaches its maximum intensity. The temporal progression from the Onset frame to the Apex frame delineates the dynamic spectrum of facial micro-expression, commencing from a state of composure and culminating at the zenith of intensity. Consequently, the discrepancy between these two frames can be utilised to discern the temporal characteristics of micro-expression movement, i.e., the change information. This can be expressed as follows:
\begin{equation}
f_{t} = f_{apex} - f_{onset}
\end{equation}
Where $f_{t}$ denotes the differential characteristics of the micro-expression motion from the calm rise to the maximum intensity phase, and $f_{onset}$, $f_{apex}$ denote the Onset and Apex frames, respectively.
Retnet has been demonstrated to be an effective architectural solution for language modelling\cite{RMT}. The present work proposes a retention mechanism for sequence modelling, which is employed to introduce temporal decay into sequence processing, thereby facilitating the understanding of temporal relationships in the model. The retention mechanism can be represented as follows:
\begin{equation}
o_n = \sum_{m=1}^{n} \gamma^{n-m} (Q_n e^{i n \theta})(K_m e^{i m \theta})^\dagger v_m
\end{equation}

The discrepancy features that have been obtained cannot be input directly into the temporal sub-branches. Consequently, it is necessary to input these features into the Conv Stem in order to obtain the token.This operation can be represented as follows: 
\begin{equation}
y_{\text{conv}} = \sum_{i=1}^{C} \sum_{j=-1}^{1} \sum_{k=-1}^{1} K_{d, i, j, k} \, x_{b, i, h+j, w+k} + b_{d}
\end{equation}
\begin{equation}
y_{\text{bn}} = \gamma_d \frac{y_{\text{conv}, b, h, w, d} - \mu_d}{\sqrt{\sigma_d^2 + \epsilon}} + \beta_d
\end{equation}
\begin{equation}
y_{\text{gelu}} = y_{\text{bn}, b, h, w, d} \Phi(y_{\text{bn}, b, h, w, d})
\end{equation}

Where x denotes the input feature, K is the convolution kernel, b is the bias term, $y_{conv}$ is the output feature map after convolution operation, $y_{bn}$ denotes the output feature map after batch normalization operation, $y_{gelu}$ is the output feature map after GELU activation function.

Subsequently, the obtained token is fed into RetNetBlock to obtain the temporal features of dimension 14 $\times$ 14 $\times$ 512.

\subsection{Spatial sub-branch}

Differential features of Apex frames and Onset frames in micro-expression recognition have been shown to provide rich information about facial muscle movements. However, it is difficult to understand the correspondence between the location where the facial muscle movement occurs and specific regions of the face with only temporal sub-branches. Spatial features, as an important part of facial information, can be further subdivided into appearance features and geometric features. Appearance features mainly present the contour information of the image, while geometric features contain geometric properties such as measurement distance and curvature.

To solve the above problem, we introduce spatial sub-branching to extract geometric features from spatial features. Given that the Onset frame serves as the starting moment of the facial micro-expression movement, when the face is in a calm state and the facial muscles have not undergone any deformation, which can provide more stable basic information about the face, we use the Onset frame as the input feature for spatial sub-branching. In order to obtain a valid spatial token, we first sliced the input features, the specific operation can be expressed as follows:
\begin{equation}
y_{b, d, h', w'} = \sum_{i=1}^{C} \sum_{j=0}^{P-1} \sum_{k=0}^{P-1} K_{d, i, j, k} \, x_{b, i, h' \cdot P + j, w' \cdot P + k} + b_{d}
\end{equation}

\begin{equation}
y_{\text{flat}} = y_{b, d, h', w'} \rightarrow y_{\text{flat}, b, h' \cdot w', d}
\end{equation}

where x is the input feature, P is the edge length of the patch, h' and w' are the indexes of the output feature graph indicating the position of the patch.

Subsequently, position information is added to each chunk. The specific operation can be represented as follows:
\begin{equation}
y_{b, n, d} = y_{b, n, d} + \text{pos\_embed}_{1, n, d}
\end{equation}

where b, n and d denote the batch size, the number of patches and embedding dimension, respectively, pos\_embed denotes a set of learnable position tensors.

Subsequently, the processed features are fed into the Transformer block to obtain spatial features. It is worth noting that we only used two layers of Transformer block to avoid extracting the identity information that is not related to the micro-expression motion. It has been shown that the LayerNorm layer can introduce strong nonlinearities and that such nonlinearities can enhance the model's representational capabilities\cite{Zhu2025DyT}. In addition, this compression behavior reflects the saturation property of biological neurons for large inputs, a phenomenon that has been observed about a century ago. To introduce this strong nonlinearity, we add a LayerNorm layer before deforming the obtained tensor to a dimension size of 14 $\times$ 14 $\times$ 512 to enhance the characterization of spatial sub-branches.

\subsection{Temporal-spatio feature fusion}

The duration of the micro-expression movement is very brief, only 1/25 to 1/2 of a second. And facial muscle movements are often strongly correlated with specific facial regions. For example, happy movements are mainly characterized by lifting of the cheeks and pulling of the corners of the mouth. Therefore, in order to effectively integrate temporal and spatial information, we propose a temporal-spatio fusion model. As shown in (d) in Figure \ref{fig2}, the third-order tensor $\mathcal{F} \in \mathbb{R}^{B \times N \times C}$ obtained in the spatial sub-branch is reshaped into $\mathcal{F} \in \mathbb{R}^{B \times C \times H \times W}$, where B stands for the batch size, N stands for the length of the sequence, C stands for the number of channels, H stands for the height of the feature, and W stands for the width of the feature. After that, it is element-wise summed with the temporal features extracted in the temporal sub-branch, which can be represented as follows:
\begin{equation}
\mathcal{F_\text{t-s}} = \mathcal{F}_{\text{t}} + \mathcal{F}_{\text{s}}
\end{equation}
Where $\mathcal{F}_{\text{t}}$ is the temporal feature, $\mathcal{F}_{\text{s}}$ is the spatial feature, and $\mathcal{F_\text{t-s}}$ is the temporal-spatio feature.

It is worth noting that it has been pointed out that the feature distribution of the model with a fully connected layer will be more extensive, not just focusing on the central object of the image\cite{zhang2017fc}. So we use a 1024 fully connected layer to extract the fused temporal-spatio features in order to obtain a broader feature distribution. The obtained temporal-spatio features are then subjected to normalization, activation and pooling operations, which can be represented as follows:

\begin{equation}
x_{\text{norm}} = \gamma \left( \frac{x - \mu}{\sqrt{\sigma^2 + \epsilon}} \right) + \beta
\end{equation}

\begin{equation}
x_{\text{swish}} = x_{\text{norm}} \cdot \frac{1}{1 + e^{-x_{\text{norm}}}}
\end{equation}

\begin{equation}
x_{\text{avg}} = \frac{1}{N} \sum_{i=1}^{N} x_{\text{swish}}^{(i)}
\end{equation}

Finally, the cross-entropy loss function is employed as the loss function in the proposed framework, which can be represented as follows:

\begin{equation}
\text{L} = -\sum_{i=1}^{N} \log \left( \frac{e^{x_{i,y_i}}}{\sum_{k=1}^{C} e^{x_{i,k}}} \right)
\end{equation}

\section{Experiments}

In this section, a series of experiments is conducted to evaluate the effectiveness of the proposed method. Firstly, the dataset, baseline and environment configurations are presented. Subsequently, a comparison is made between TSFmicro and state-of-the-art methods, thereby demonstrating the advancement of TSFmicro. Furthermore, an analysis of different fusion modes was conducted to ascertain the impact of these modes on the performance of micro-expression recognition.

\subsection{Experimental Setup}

\subsubsection{Datasets} 
To validate the effectiveness of TSFmicro, we conducted experiments on the following representative micro-expression datasets:

\textbf{CASME \uppercase\expandafter{\romannumeral2}} (2014)\cite{casme2}.The 7 emotional categories are labelled, but in this paper we only use the 5 categories (Happiness, Disgust, Repression, Surprise and Other) and the 3 categories under the MGC2019 protocol (Negative, Positive and Surprise).

\textbf{SAMM} (2018)\cite{samm}.The 8 emotion categories were labelled in SAMM, and in this paper we only use the 5-category (Happiness, Anger, Contempt, Surprise and Other) and the 3-category under the MGC2019 protocol.

\textbf{CAS(ME)$^3$} (2022)\cite{casme3}. The current micro-expression dataset and 7 emotion categories are labelled. In our experiments, 7 categories and 4 categories (Negative, Positive, Surprise and Others) are used for classification tasks.

In accordance with the majority of the studies, the leave-one-subject-out (LOSO) cross-validation method is employed. The ACCuracy (ACC), Unweighted F1 score (UF1) and unweighted average recall (UAR) are used as the evaluation metrics:
\begin{equation}
\text{UF1} = \frac{1}{C} \sum_{i=1}^{C} \frac{2 \times \text{TP}_i}{\text{TP}_i + \text{FP}_i + \text{FN}_i}
\end{equation}
\begin{equation}
\text{UAR} = \frac{1}{C} \sum_{i=1}^{C} \frac{\text{TP}_i}{N_i}
\end{equation}

\begin{table*}[h]
\caption{Detailed information about the dataset for the class=5/7 condition. The abbreviations Exp, Num, Dis, Hap, Sur, Rep, Ang, Con, Oth denote Expression, Sample Size, Disgust, Happiness, Surprise, Depression, Anger, Contempt and Others.}
\centering
\begin{tabular}{@{}c|c|c|c|c|c}
\toprule
\multicolumn{2}{c|}{CASME \uppercase\expandafter{\romannumeral2}} & \multicolumn{2}{c|}{SAMM} &\multicolumn{2}{c}{CAS(ME)$^3$} \\
Exp & Num & Exp & Num & Exp & Num\\
\midrule
Hap & 32 & Hap & 26 & Hap & 55 \\
Dis & 62 & Ang & 57 & Ang & 64 \\
Rep & 27 & Con & 12 & Dis & 250 \\
Sur & 28 & Sur & 15 & Fea & 86 \\
Oth & 99 & Oth & 26 & Sad & 57 \\
-&-&-&-& Sur & 184 \\
-&-&-&-& Oth & 159 \\
\midrule
Total & 248 & Total & 136 & Total & 855 \\
\midrule
\end{tabular}
\label{tab1}
\end{table*}

\begin{table*}[h]
\caption{Detailed information about the dataset for the class=3/4 condition. The abbreviations Exp, Num, Neg, Pos, Oth denote Expression, Sample Size, Negative, Positive and Others.}
\centering
\begin{tabular}{@{}c|c|c|c|c|c}
\toprule
\multicolumn{2}{c|}{CASME \uppercase\expandafter{\romannumeral2}} & \multicolumn{2}{c|}{SAMM} &\multicolumn{2}{c}{CAS(ME)$^3$} \\
Exp & Num & Exp & Num & Exp & Num\\
\midrule
Neg & 95 & Neg & 92 & Neg & 457 \\
Pos & 32 & Pos & 26 & Pos & 55 \\
Sur & 28 & Sur & 15 & Sur & 184 \\
-&-&-&-& Oth & 159 \\
\midrule
Total & 155 & Total & 133 & Total & 855 \\
\midrule
\end{tabular}
\label{tab2}
\end{table*}

\subsubsection{Baselines}  
We compare our proposed method with the following baselines:
\begin{itemize}
    \item AU-GCN \cite{AUGCN}: Action unit graph convolutional network, which utilizes conditional probability-based adjacency matrix and node embedding, effectively captures the correlation between action units (AUs). It improves micro-expression recognition by integrating the action unit information into the facial graph representation through a two-channel fusion mechanism.
    \item MMNet \cite{li2022mmnet}: Muscle motion-guided network that focuses on modeling local subtle muscle motion patterns through a continuous attention (CA) block and incorporates facial position information via a position calibration (PC) module based on vision transformers. It extracts motion-pattern features and facial position embeddings through two branches, then fuses them for micro-expression recognition.
    \item C3DBed \cite{C3DBed}: A micro-expression recognition model that combines a three-dimensional convolutional neural network (C3D) and a transformer model. It extracts local features from micro-expression images by learning attention weights, addressing the low-intensity and information redundancy challenges in MER. The model embeds apex frame patches into latent features using C3D, fuses these with position embeddings, and processes them through a transformer encoder to learn attention weights. Joint loss optimization is used to improve model performance.
    % \item $\mu$-BERT \cite{}:A micro-expression recognition model based on BERT. It uses Diagonal Micro-Attention (DMA) to detect subtle differences between consecutive frames and a Patch of Interest (PoI) module to focus on relevant facial regions while ignoring background noise. 
    \item DecFlow \cite{DecFlow}: A decomposed facial flow model that addresses the challenges of non-rigid motion and entangled representation in facial optical flow estimation. It introduces a facial semantic-aware encoder and a decomposed flow decoder to accurately estimate and separate facial optical flow into head and expression components, achieving significant improvements in facial flow estimation accuracy and enhancing micro-expression recognition performance.
    % \item TleMer \cite{}: A micro-expression recognition model that combines transfer learning, dense connections, and mixed attention. It uses Euler Video Magnification to enhance micro-expression intensity, a densely connected feature extraction module to reduce feature loss, and a mixed attention mechanism to focus on discriminative features. This approach effectively addresses the challenges of limited training data and low expression intensity in micro-expression recognition.
    \item MPFNet \cite{ma2025}: A Multi-Prior Fusion Network for video-based micro-expression recognition that employs a progressive training strategy. It uses a dual-stream architecture with a Triplet Network and motion-amplified ME dataset to capture generic and advanced features. The model integrates two pre-trained models within a meta-learning framework for classification, utilizing I3D with Coordinate Attention blocks to enhance spatiotemporal and channel feature learning.
    \item LP-GCN \cite{zhang2025}:A approach for micro-expression recognition that integrates facial 3D structure and motion features using a lightweight point cloud network and graph convolutional network. It represents facial motion in 3D space, segments the face into eight regions, and extracts motion features from each region. 
\end{itemize}
For baselines, we use the source codes provided by the authors and tune hyper-parameters on the validation sets to get their best performances.

\subsubsection{Implementation Details}
In the course of our experiments, we employed the MTCNN algorithm to identify the salient feature points in the original datasets. These feature points were then used to guide the cropping of the face in each image. Subsequently, all images are resized to 224 $\times$ 224 pixels. To prevent overfitting, we employ techniques such as horizontal flipping and random cropping to enhance the dataset. With regard to the model architecture, a five-layer RetNet network was employed to form the temporal branch, while a ViT was used to construct the spatial branch. In the training phase, the AdamW optimiser is employed to optimise the model. The initial learning rate is set to 0.0008, the batch size is set to 32, and the loss function utilises the cross-entropy loss function. The learning rate declines exponentially over the course of 50 epochs. All experiments were conducted on a single NVIDIA RTX 4060 Ti GPU, utilising the PyTorch framework.

\begin{figure*}[!h]
\centering
\includegraphics[width=1\textwidth]{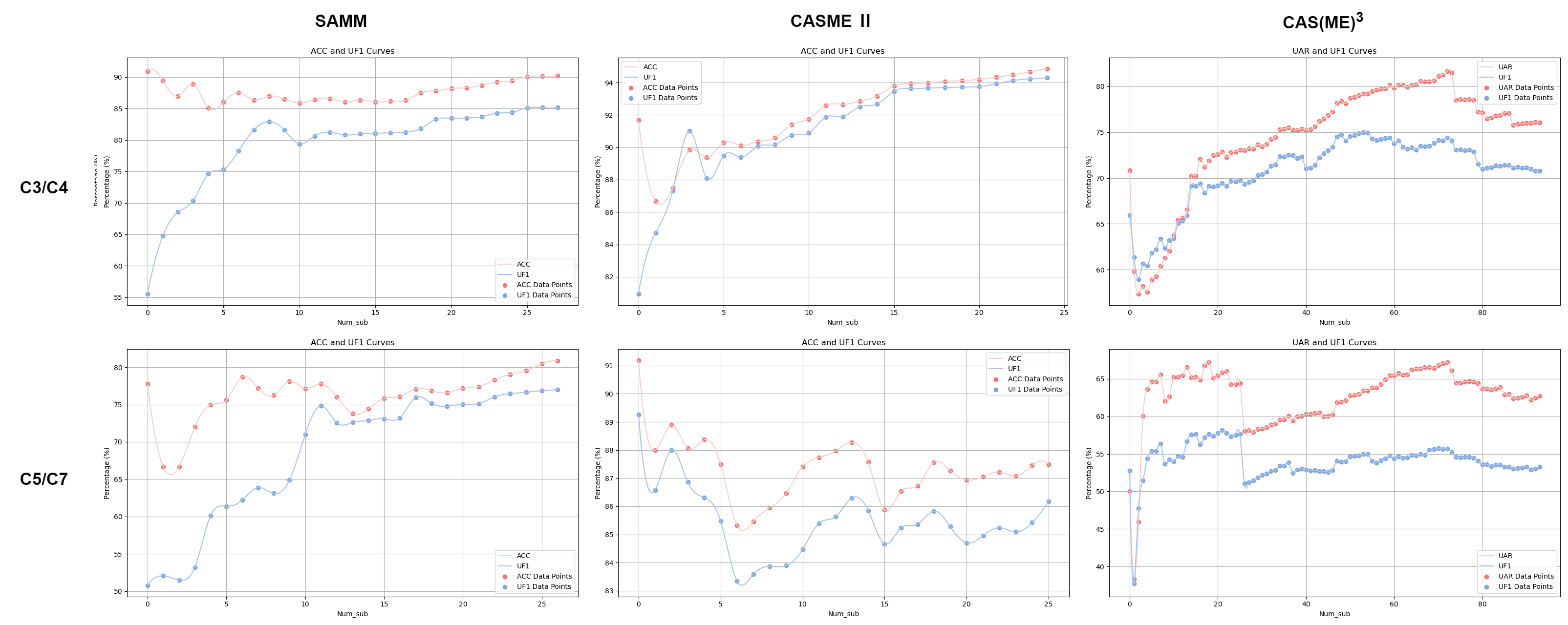}
\caption{Evaluation scores of SAMM, CASME \uppercase\expandafter{\romannumeral2} and CAS(ME)$^3$ datasets under 3/4-class and 5/7-class classification conditions during TSFmicro training.}
\label{fig5}
\end{figure*}

\begin{figure*}[!h]
\centering
\includegraphics[width=1\textwidth]{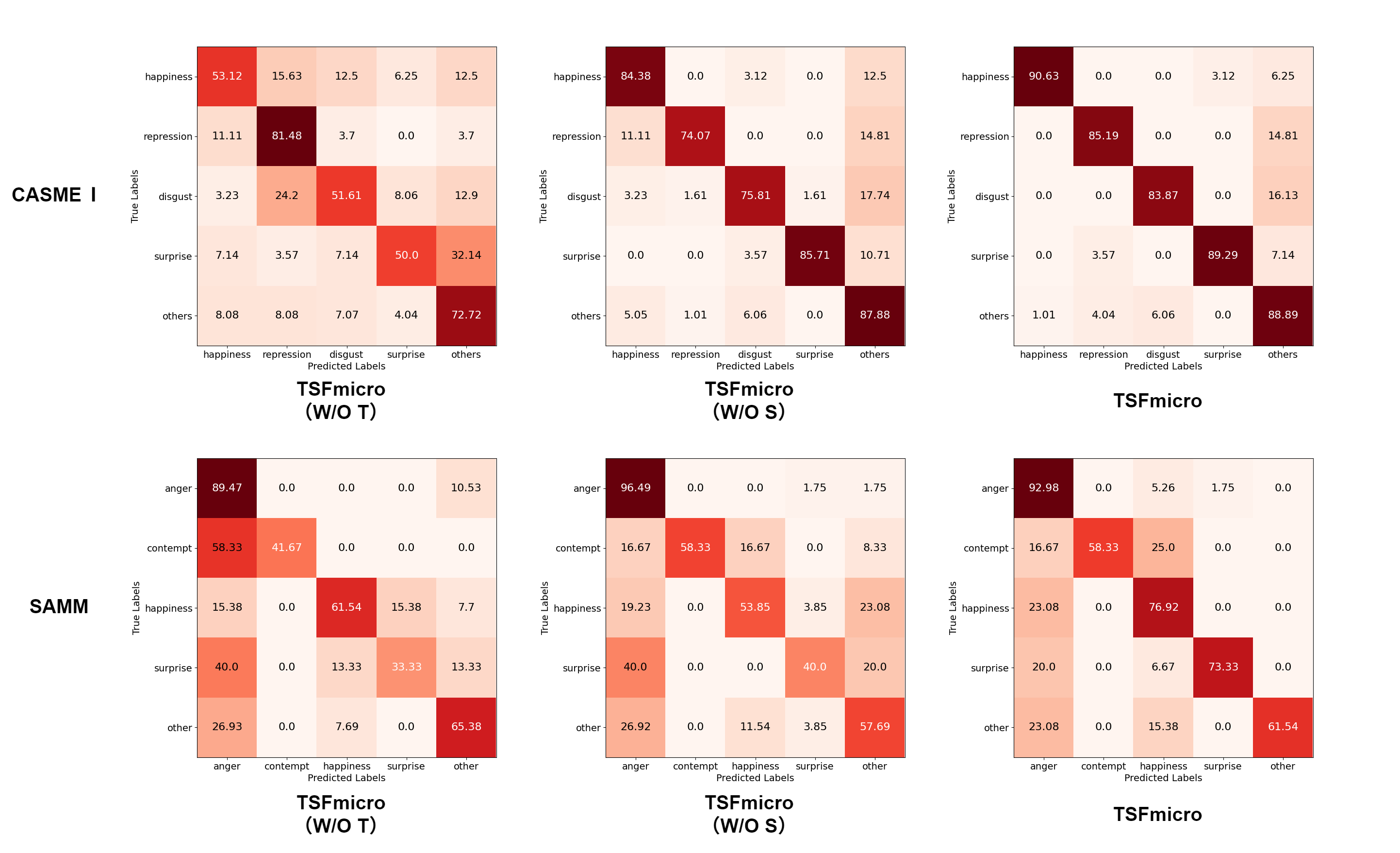}
\caption{Confusion matrix evaluation results of our proposed TSFmicro framework with different datasets.}
\label{fig3}
\end{figure*}

\begin{table}[t]
\caption{Performance comparison on CAS(ME)$^3$.}
\centering
\begin{tabular}{@{}c|cc|cc}
\toprule
\multirow{2}{*}{Model} & \multicolumn{2}{c|}{7-class(\% ↑)} & \multicolumn{2}{c}{4-class(\% ↑)} \\
& UF1 & UAR & UF1 & UAR \\
\midrule
AlexNet\cite{casme3} '22& 17.59 & 18.01 & 29.15 & 29.10 \\
% $\mu$-BERT\cite{u-BERT} '23& 32.64 & 32.54 & 47.18 & 49.13 \\
SFAMNet\cite{SFAMNet} '24& 23.65 & 23.73 & 44.62 & 47.97 \\
ATM-GCN\cite{ATM-GCN} '24& \textcolor{blue}{43.08} & \textcolor{blue}{42.83} & 54.23 & 53.49 \\
MPACNet\cite{zhao2025} '25& 42.01 & 42.44 & \textcolor{blue}{58.14} & \textcolor{blue}{56.40} \\ 
LP-GCN\cite{zhang2025} '25& 35.64 & 41.59 & 47.64 & 53.66 \\
\midrule
TSFmicro(Ours) & \textcolor{red}{53.29} & \textcolor{red}{62.73} & \textcolor{red}{70.75} & \textcolor{red}{76.03}\\
\bottomrule
\end{tabular}
\label{tab3}
\end{table}

\begin{table}[!tbp]
\caption{Performance comparison on CASME \uppercase\expandafter{\romannumeral2}.}
\centering
\begin{tabular}{@{}c|cc|cc}
\toprule
\multirow{2}{*}{Model} & \multicolumn{2}{c|}{5-class(\% ↑)} & \multicolumn{2}{c}{3-class(\% ↑)} \\
& ACC & UF1 & ACC & UF1 \\
\midrule
MERSiamC3D\cite{MERSiamC3D} '21 & 81.89 & 83.00 & 87.63 & 88.18 \\
AU-GCN\cite{AUGCN} '21 & 74.27 & 70.47 & 87.10 & 87.98\\
MMNet\cite{li2022mmnet} '22& \textcolor{blue}{86.09} & \textcolor{blue}{85.21} & 92.25 & \textcolor{blue}{93.68} \\
AMAN\cite{AMAN} '22 & 75.40 & 71.25 & - & -\\
FRL-DGT\cite{FRL} '23 & 75.70 & 74.80 & 88.50 & 89.80\\
C3DBed\cite{C3DBed} '23 & 77.64 & 75.20 & 88.82 & 89.78 \\
% $\mu$-BERT\cite{u-BERT} '23 & 83.48 & \textcolor{blue}{85.53} & 89.14 & 90.34 \\
ATM-GCN\cite{ATM-GCN} '24& - & - & 90.42 & 90.48 \\
DecFlow\cite{DecFlow} '24 & 82.10 & 81.80 & \textcolor{blue}{94.20} & 93.10 \\
% TleMer\cite{TleMer} '24 & \textcolor{blue}{86.35} & 84.08 & \textcolor{blue}{94.67} & 93.40 \\
DSSTDN\cite{nikin2025} '25 & 77.91 & 77.55 & 87.82 & 87.63 \\
MPFNet\cite{ma2025} '25 & 82.00 & 81.90 & 89.70 & 89.80 \\
\midrule
TSFmicro(Ours) & \textcolor{red}{87.50} & \textcolor{red}{86.17} & \textcolor{red}{94.84} & \textcolor{red}{94.30}\\
\bottomrule
\end{tabular}
\label{tab4}
\end{table}

\begin{table}[!t]
\caption{Performance comparison on SAMM.}
\centering
\begin{tabular}{@{}c|cc|cc}
\toprule
\multirow{2}{*}{Model} & \multicolumn{2}{c|}{5-class(\% ↑)} & \multicolumn{2}{c}{3-class(\% ↑)} \\
& ACC & UF1 & ACC & UF1 \\
\midrule
MERSiamC3D\cite{MERSiamC3D} '21 & 68.75 & 64.00 & 72.80 & 74.75\\
AU-GCN\cite{AUGCN} '21 & 74.26 & 70.45 & 78.90 & 77.51\\
MMNet\cite{li2022mmnet} '22& \textcolor{blue}{80.14} & 72.91 & \textcolor{blue}{90.22} & 83.91 \\
AMAN\cite{AMAN} '22 & 68.85 & 66.82 & - & - \\
C3DBed\cite{C3DBed} '23 &  75.73 & 72.16 & 80.67 & 81.26 \\
% $\mu$-BERT†\cite{u-BERT} '23 & \textcolor{red}{84.75} & \textcolor{red}{83.86} & - & - \\
SRMCL\cite{SRMCL} '24& 74.63& 65.99& 88.66& 84.70\\
% TleMer†\cite{TleMer} '24 & 84.44 & 81.78 & 90.15 & 84.85 \\
DSSTDN\cite{nikin2025} '25 & 78.68 & \textcolor{blue}{76.41} & 86.36 & \textcolor{blue}{85.17} \\
MPFNet\cite{ma2025} '25 & 70.40 & 69.5 & 82.80 & 82.20 \\
\midrule
TSFmicro(Ours) & \textcolor{red}{80.88} & \textcolor{red}{77.00} & \textcolor{red}{90.97} & \textcolor{red}{85.67}\\
\bottomrule
% \multicolumn{4}{l}{† Additional pre-training phase required.}
\end{tabular}
\label{tab5}
\end{table}

\subsection{Main Results}

Table \ref{tab3} illustrates the efficacy of our TSFmicro on the CAS(ME)$^3$ dataset in comparison to alternative methodologies employed over the past three years. Our TSFmicro demonstrates excellent performance in both the 7-categorisation and 4-categorisation tasks, significantly outperforming the other methods and achieving the best results. In particular, with regard to the 7-categorisation task, the TSFmicro model exhibits a superior performance compared to the second-ranked ATM-GCN approach, with an improvement of 10.21\% and 19.9\% points on the UF1 and UAR metrics, respectively. This advantage is also evident in the 4-categorisation task, with respective values of 16.52 and 22.54\%. Furthermore, the model was tested on smaller datasets, namely CASME \uppercase\expandafter{\romannumeral2} and SAMM. The results are presented in Table \ref{tab4} and \ref{tab5}. In the 5-categorisation task on the CASME \uppercase\expandafter{\romannumeral2} dataset, TSFmicro achieves the best performance, with an improvement of 1.41\% points in ACC and 0.96\% points in UF1 over the second-ranked MMNet method. Furthermore, the improvement reached 0.64 and 0.62\% points in the 3-categorization task, respectively. The MMNet result was obtained by replicating the original author's steps on a Nvidia 4060 Ti GPU.

In addition, we curve-fitted the scores during the TSFmicro training process, and the results are shown in Figure \ref{fig5}. Influenced by cross-cultural factors, the samples from different cultures differ in the expression and suppression of micro-expressions, which makes the recognition of the model more challenging\cite{li2025nc}. In contrast, the samples of CASME \uppercase\expandafter{\romannumeral2} and CAS(ME)$^3$ datasets, both of which originate from the same cultural background, have higher consistency in their micro-expression expression paradigms, and thus the metrics changes of these two datasets show more consistent characteristics during the training process.

\begin{table*}[!h]
\caption{Ablation study on CASME \uppercase\expandafter{\romannumeral2} and SAMM.}
\centering
\begin{tabular}{@{}c|cc|cc|cc}
\toprule
\multirow{2}{*}{DataSet}& \multicolumn{2}{c|}{Model}& \multicolumn{2}{c|}{5-class(\% ↑)}&\multicolumn{2}{c}{3-class(\% ↑)} \\ 
 & Temporal & Spatial & ACC & UF1 & ACC & UF1 \\
\midrule
\multirow{3}{*}{CASME \uppercase\expandafter{\romannumeral2}}
& $\times$ & $\checkmark$ & 63.30 & 55.19 & 81.29 & 75.18 \\
& $\checkmark$ & $\times$ & \textcolor{blue}{83.87(+20.57)} & \textcolor{blue}{81.05(+25.86)} & \textcolor{blue}{93.55(+12.26)} & \textcolor{blue}{90.23(+15.05)} \\
& $\checkmark$ & $\checkmark$ & \textcolor{red}{87.50(+24.20)} & \textcolor{red}{86.17(+30.98)} & \textcolor{red}{94.84(+13.55)} & \textcolor{red}{94.30(+19.12)} \\
\midrule
\multirow{3}{*}{SAMM}
& $\times$ & $\checkmark$ & 69.12 & 50.96 & 79.69 & 64.50 \\
& $\checkmark$ & $\times$ & \textcolor{blue}{75.00(+5.88)} & \textcolor{blue}{65.39(+14.43)} & \textcolor{blue}{89.47(+9.78)} & \textcolor{blue}{83.38(+18.88)} \\
& $\checkmark$ & $\checkmark$ & \textcolor{red}{80.88(+11.76)} & \textcolor{red}{77.00(+26.04)} & \textcolor{red}{90.97(+10.54)} & \textcolor{red}{85.67(+20.66)} \\
\midrule
\end{tabular}
\label{tab6}
\end{table*}

\begin{table*}[!t]
\caption{Fusion study on CASME \uppercase\expandafter{\romannumeral2} and SAMM.}
\centering
\begin{tabular}{@{}c|c|cc|cc}
\toprule
\multirow{2}{*}{DataSet}& \multirow{2}{*}{Fusion mode}& \multicolumn{2}{c|}{5-class(\% ↑)}&\multicolumn{2}{c}{3-class(\% ↑)} \\ 
 & & ACC & UF1 & ACC & UF1 \\
\midrule
\multirow{4}{*}{CASME \uppercase\expandafter{\romannumeral2}}
& T-S(early) & 82.25 & 79.92 & 90.96 & 89.33 \\
& T to S & \textcolor{blue}{85.88(+3.63)} & \textcolor{blue}{84.78(+4.86)} & \textcolor{blue}{94.19(+3.23)} & \textcolor{blue}{93.53(+4.20)} \\
& S to T & 84.27(+2.02) & 83.35(+3.43) & 93.55(+2.59) & 92.75(+3.42) \\
& T-S(late) & \textcolor{red}{87.50(+5.25)} & \textcolor{red}{86.17(+6.25)} & \textcolor{red}{94.84(+3.88)} & \textcolor{red}{94.30(+4.97)} \\
\midrule
\multirow{4}{*}{SAMM}
& T-S(early) & 73.52 & 62.92 & 84.21 & 72.60 \\
& T to S & 77.21(+3.69) & 67.34(+4.42) & 88.72(+4.51) & 79.35(+6.75) \\
& S to T & \textcolor{blue}{79.41(+5.89)} & \textcolor{blue}{68.85(+5.93)} & \textcolor{blue}{89.47(+5.26)} & \textcolor{blue}{80.30(+7.70)} \\
& T-S(late) & \textcolor{red}{80.88(+7.36)} & \textcolor{red}{77.00(+14.08)} & \textcolor{red}{90.97(+6.02)} & \textcolor{red}{85.67(+12.56)} \\
\midrule
\end{tabular}
\label{tab7}
\end{table*}

\subsection{Ablation Study}
In order to validate the effectiveness of each of the proposed components, three variants were designed: TSFmicro (W/O T), TSFmicro (W/O S) and the complete TSFmicro. The effects of these variants were then compared. The findings of the aforementioned studies are presented in Tables \ref{tab6} and \ref{tab8}.

As demonstrated in the tables, the elimination of the temporal sub-branch of TSFmicro(W/O T) resulted in ACC and UF1 attaining 63.30 and 55.19 for the five classification tasks on the CASME \uppercase\expandafter{\romannumeral2} dataset, while on the SAMM dataset, ACC and UF1 attained 69.12\% and 50.96\%, respectively. This finding suggests that while the spatial information extracted from the spatial sub-branch demonstrates the capacity for micro-expression recognition, its static characteristics appear to impose limitations on the enhancement of model performance. In comparison with TSFmicro(W/O T), TSFmicro(W/O S) with the elimination of the spatial sub-branch enhances the performance of the 5-categorisation task by 20.57\% (ACC) and 25.86\% (UF1) on the CASME \uppercase\expandafter{\romannumeral2} dataset, which are 5.88\% (ACC) and 14.43\% (UF1) on the SAMM dataset. This finding suggests that the temporal sub-branch contains the majority of the information necessary for micro-expression recognition. However, when relying solely on temporal information, the model is incapable of accurately localising the micro-expression action at the moment of occurrence. As demonstrated in the confusion matrix depicted in Figure \ref{fig3}, the TSFmicro(W/O S) model demonstrates a high level of performance in categories characterised by a substantial sample size. However, in categories comprising a limited number of samples, such as negative emotions such as repression and disgust, the model exhibits a tendency to erroneously categorise these samples as others, which are typically associated with a greater sample size.

Finally, the TSFmicro with complete inclusion of spatial and temporal sub-branches achieved the highest performance, reaching 87.50\% (ACC) and 86.17\% (UF1) on the CASME \uppercase\expandafter{\romannumeral2} dataset. On the SAMM dataset, it achieved 80.88\% (ACC) and 77.00\% (UF1).The enhancement in UF1 scores indicates that the fusion of spatio-temporal information can effectively enhance the model's classification capability in categories with limited sample sizes. This further demonstrates that spatial information can complement the static properties in temporal information, thus enriching the feature space to counteract the negative effects of smaller sample sizes.

\begin{figure*}[!h]
\centering
\includegraphics[width=1\textwidth]{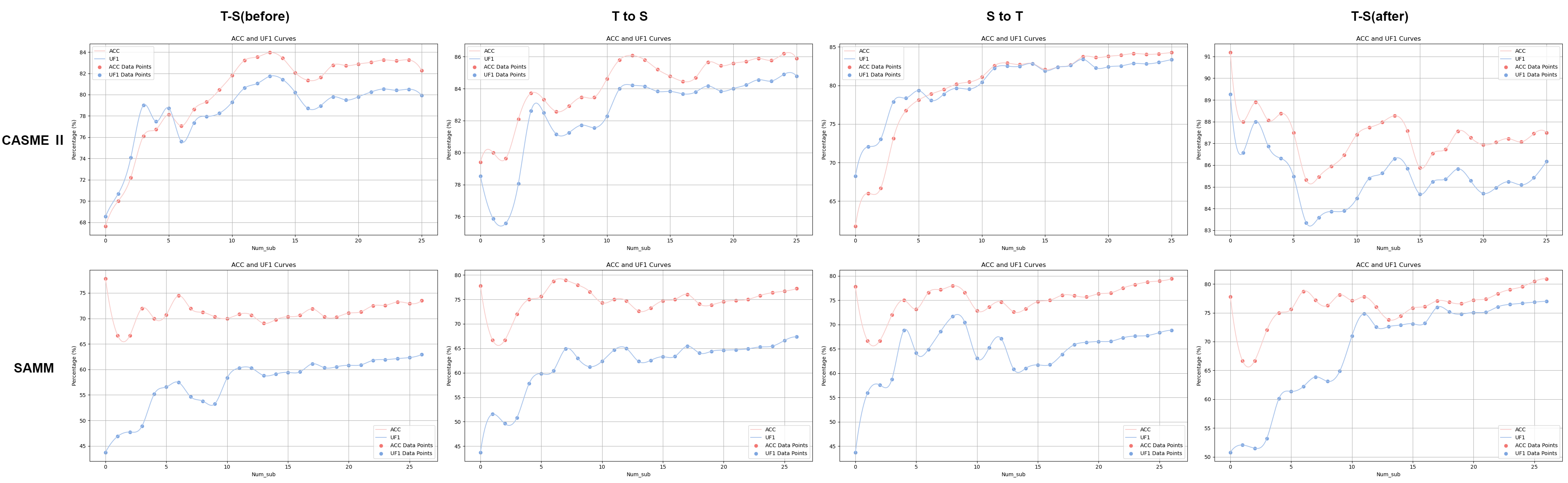}
\caption{Evaluation scores of different fusion methods on the SAMM and CASME  \uppercase\expandafter{\romannumeral2} datasets under 5 classification conditions during TSFmicro training.}
\label{fig6}
\end{figure*}

\subsection{Fusion Study}
In order to further explore the effect of fusion modality on micro-expression recognition performance, we propose four fusion strategies from the perspective of modal fusion, namely T-S (early), T to S, S to T and T-S (late). The experiments are conducted on the CASME \uppercase\expandafter{\romannumeral2} and SAMM datasets, with the specific experimental results displayed in Tables \ref{tab7} and \ref{tab8}. Furthermore, we implemented curve fitting of the scores during the training process, as demonstrated in Figure \ref{fig6}. This figure provides a visual representation of the impact of diverse fusion strategies on the model training process.

T-S (early) is a pre-fusion strategy that fuses temporal and spatial information prior to the recognition backbone following a simple convolutional extraction. The fused features are then provided as a whole input to the recognition backbone for classification. However, the performance of this fusion strategy is lower than that of TSFmicro (W/O S). The findings indicate a 3.63\% and 4.86\% lag for ACC and UF1, respectively, in the five classification task in the CASME  \uppercase\expandafter{\romannumeral2} dataset, and a 3.69\% and 4.42\% lag for ACC and UF1, respectively, in the SAMM dataset. This may be due to the fact that the temporal modelling in the temporal information is interfered by the static features in the spatial information, resulting in the loss of some of the spatio-temporal information, which affects the performance of the model.

T to S and S to T use sequential fusion strategy. t to S first extracts temporal features, then fuses them with spatial information and inputs spatial sub-branches to complete the classification. In contrast, S to T extracts spatial features first, then fuses them with temporal information and inputs temporal sub-branches for classification. Both fusion methods perform better than TSFmicro (W/O S), which indicates that spatial information can effectively enhance the characterization of temporal information.

Finally, the T-S (late) fusion strategy demonstrates optimal performance. This strategy extracts spatio-temporal features using spatial and temporal sub-branches, respectively, in order to preserve the temporal relationships and geometric features in the spatio-temporal information to a greater extent. The temporal sub-branch focuses on learning the temporal relationship when micro-expressions occur, while the spatial sub-branch learns the geometric features of the face. The two-branch structure is effective in avoiding spatio-temporal interference with each other. Furthermore, the fusion occurs in the high-level feature space to form a complementary "where-how" relationship at the semantic level. This provides richer semantic information for the model, thereby enhancing its classification ability.

\begin{figure*}[!h]
\centering
\includegraphics[width=1\textwidth]{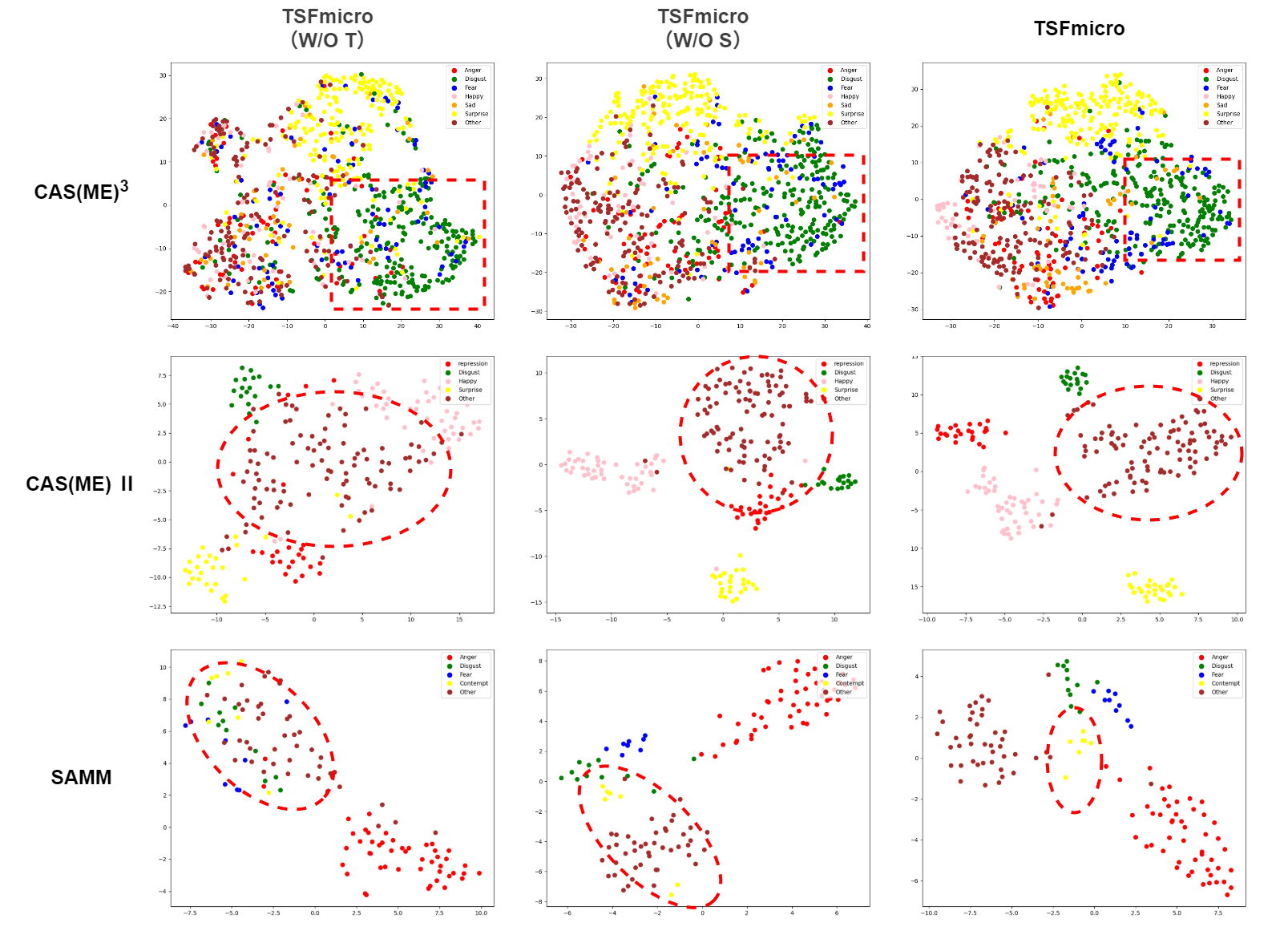}
\caption{The feature distributions of the three TSFmicro variants are visualized for three different datasets.}
\label{fig4}
\end{figure*}

\subsection{Visualization Analysis}
\textbf{Visualization of Feature Distribution.}To more intuitively analyze the influence of spatial and temporal sub-branches on the overall framework, we used the t-SNE\cite{tsne} technique to visualize the distribution of micro-expression features extracted from the model. Figure \ref{fig4} illustrates the comparison of the feature distributions of the three variants under the three datasets. In particular, the first column shows the effect of removing the temporal sub-branch and the second column shows the effect of removing the spatial sub-branch. Compared to the spatial sub-branch, the temporal sub-branch clusters samples of the same emotion category more compactly. This suggests that the temporal information extracted by the temporal sub-branch can support most of the information required for micro-expression recognition. However, for similar emotion categories, such as negative emotions like FEAR and DISGUST, the decision boundary still exhibits a certain degree of overlap. This suggests that the model may encounter difficulties in distinguishing these similar emotions when relying only on temporal sub-branches. In contrast, the decision boundary of the full TSFmicro is more emotional and the ability to distinguish between different emotion categories is enhanced. This suggests that the model can effectively spatial and temporal information, thus improving the performance of recognition.

\begin{figure*}[!h]
\centering
\includegraphics[width=1\textwidth]{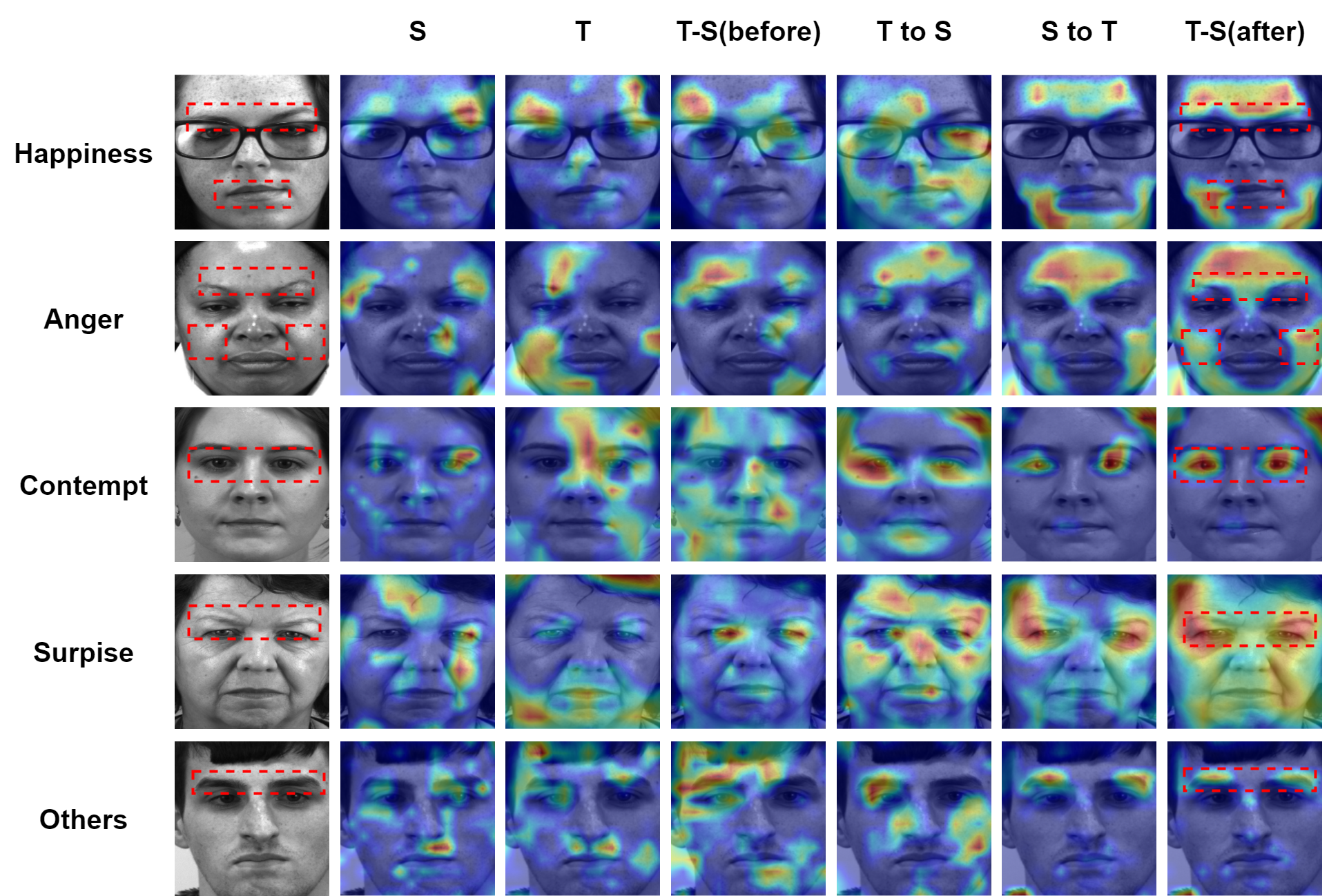}
\caption{The visualization of feature heatmaps corresponds to the different categories in the SAMM dataset.}
\label{fig7}
\end{figure*}

\textbf{Visualization of Feature Heatmaps.}To better understand the impact of different fusion strategies on micro-expression recognition performance. We use Grad-CAM\cite{gradcam} to visualise the activation heatmap. As shown in Figure \ref{fig7}, we show the heatmaps of different fusion strategies under 5 classification conditions. Also, we provide the original images of the samples as a reference. From Figure \ref{fig7}, we can see that the T-S (early) heatmap is relatively cluttered and cannot focus on the region where the action occurs, e.g., the heatmap of the feature of anger can only focus on the eyebrow and the corner of the mouth. Comparing T to S and S to T, it can be found that although T to S can understand the temporal relationship when the action occurs, it cannot correspond well to the location when the action occurs. T to S can capture the temporal relation of eyebrow gathering, but it cannot focus on the eyebrow region. S to T can correctly map the temporal relation to the corresponding region, but it also loses part of the temporal information. As in anger the temporal information is lost for the left cheek lift. the T to S (late) heatmap performs the best and can map the temporal relationship to a specific region well while also retaining the full temporal information.

\begin{table*}[!t]
\caption{Detailed information about the fusion experiment.}
\centering
\begin{tabular}{@{}c|c|ccccc|c|c}
\toprule
\multirow{1}{*}{DataSet} & \multirow{1}{*}{Fusion mode} & Happiness & Repression & Disgust & Surpise & Others & ACC(\% ↑)& UF1(\% ↑)\\
\midrule
\multirow{6}{*}{CASME \uppercase\expandafter{\romannumeral2}}
& S & 53.12 & 81.48 & 51.61 & 50.00 & 72.72 & 63.30 & 55.19 \\
& T & \textcolor{blue}{84.38} & 74.07 & 75.81 & 85.71 & 87.88 & 83.87 & 81.05 \\
& T-S(early) & 75.00 & 70.37 & \textcolor{blue}{79.03} & \textcolor{red}{92.86} & 78.79 & 82.25 & 79.92 \\
& T to S & 78.12 & \textcolor{blue}{85.19} & 75.81 & \textcolor{blue}{89.29} & \textcolor{red}{92.93} & \textcolor{blue}{85.88} & \textcolor{blue}{84.78} \\
& S to T & 81.25 & \textcolor{red}{92.59} & 75.81 & 85.71 & 77.78 & 84.27 & 83.35 \\
& T-S(late) & \textcolor{red}{90.63} & \textcolor{blue}{85.19} & \textcolor{red}{83.87} & \textcolor{blue}{89.29} & \textcolor{blue}{88.89} & \textcolor{red}{87.50} & \textcolor{red}{86.17} \\
\midrule
& & Happiness & Anger & Contempt & Surpise & Others & ACC(\% ↑)& UF1(\% ↑)\\
\midrule
\multirow{6}{*}{SAMM}
& S & 61.54 & 89.47 & 41.67 & 33.33 & 65.38 & 69.12 & 50.96 \\
& T & 53.85 & \textcolor{blue}{96.49} & \textcolor{red}{58.33} & 40.00 & 57.69 & 75.00 & 65.39 \\
& T-S(early) & 61.54 & 91.23 & \textcolor{blue}{50.00} & \textcolor{blue}{60.00} & 65.38 & 73.52 & 62.92\\
& T to S & \textcolor{blue}{73.08} & \textcolor{red}{98.25} & 33.33 & 53.33 & 57.69 & 77.21 & 67.34 \\
& S to T & 69.23 & 94.74 & \textcolor{blue}{50.00} & 46.67 & \textcolor{blue}{73.08} & \textcolor{blue}{79.41} & \textcolor{blue}{68.85} \\
& T-S(late)  & \textcolor{red}{76.92} & 94.74 & \textcolor{red}{58.33} & \textcolor{red}{73.33} & \textcolor{red}{69.23} & \textcolor{red}{80.88} & \textcolor{red}{77.00} \\
\midrule
\end{tabular}
\label{tab8}
\end{table*}

\section{Discussion}
It is noteworthy that in the fusion study, we found that the performance of the two fusion strategies, T to S and S to T, differed on different datasets. These disparities not only mirror the attributes of the fusion strategies in question, but also elucidate the influence of dataset characteristics on model performance. Specifically, T to S outperforms S to T in the five classification tasks of the CASME \uppercase\expandafter{\romannumeral2} dataset, with an improvement of 1.61\% in the ACC index and 1.43\% in the UF1 index. Conversely, T to S lags behind S to T by 2.2\% in the ACC index and 1.51\% in the UF1 index in the five classification tasks of the SAMM dataset. This demonstrated a consistent trend across the three classification tasks.

This discrepancy may be attributable to the cross-cultural factor.The samples in the SAMM dataset encompass a broad spectrum of cultural backgrounds, while the samples in the CASME \uppercase\expandafter{\romannumeral2} dataset are predominantly Chinese. Furthermore, a comparison of the two datasets reveals that the SAMM dataset exhibits a greater age span than the CASME \uppercase\expandafter{\romannumeral2}dataset. This results in the SAMM dataset containing more complex facial morphology information. It is hypothesised that there may be significant differences in the expression and suppression of micro-expressions among people from different cultures. For instance, in Asian cultures, individuals tend to express emotions through subtle movements at the corners of their eyes or mouths. In contrast, European cultures may be more inclined to rely on overall facial expression changes. The results presented in Table \ref{tab8} further corroborate this perspective. For instance, in the 'Others' category, T to S demonstrates superior performance in comparison to S to T on the CASME \uppercase\expandafter{\romannumeral2} dataset; conversely, the opposite is observed on the SAMM dataset.
In such cases, the method that processes spatial information first (i.e., from S to T) may be more advantageous.

Furthermore, the integration of spatio-temporal information has been demonstrated to assist in addressing the category imbalance issue. The results in Table \ref{tab8} demonstrate the effectiveness of all four fusion strategies in enhancing the performance of emotion recognition for categories with a limited number of categories, such as Happiness and Surprise. This finding indicates that the complementarity of spatio-temporal information can enhance the model's sensitivity to emotions with a limited number of categories, thereby improving the robustness of the model.

% \section{Limitation}
% Despite the proposed TSFmicro framework, there are still many unsolved problems in the field of dynamic micro-expression recognition. One notable issue is the challenge of cross-cultural scenarios. For example, the samples in the SAMM dataset cover a wide range of cultural backgrounds, in which case, even though the performance of TSFmicro is optimal compared to other state-of-the-art methods, there is still a large room for improvement when comparing with the performance results in the CASME \uppercase\expandafter{\romannumeral2} dataset. This suggests that there is a generalization bottleneck in the current method to cope with different cultural expression habits and facial morphology differences, and the model may have difficulty in adapting to changes in micro-expression features due to cultural differences in cross-cultural scenarios. Another major issue is the imbalance of question categories. In the micro-expressions dataset, the number of negative emotion samples is low, which leads to a lower accuracy of the model in recognizing negative emotions. This problem is particularly prominent in the SAMM dataset, which may further limit the model's performance in real-world applications. To address these issues, future research could explore methods such as transfer learning and contrast learning to improve the model's generalization ability in cross-cultural scenarios and its ability to handle unbalanced data.

\section{Conclusion}
In this paper, we propose a novel dual-stream framework, TSFmicro, for dynamic micro-expression recognition, which improves the performance of micro-expression recognition by fusing temporal and spatial features. The proposed methodology effectively addresses the challenges posed by temporal dynamics and spatial localization in micro-expression recognition, achieving efficient feature fusion through multimodal fusion techniques. The primary innovations of TSFmicro comprise a novel spatio-temporal feature fusion framework, which captures the dynamic temporal information by calculating the difference between the start frame and the vertex frame, and combines it with the positional information extracted from spatial sub-branches to achieve efficient spatio-temporal information fusion. Position embedding is introduced in the spatial sub-branches while avoiding the extraction of identity information unrelated to micro-expression motion by reducing the number of network layers. A parallel spatio-temporal fusion method is proposed to fuse spatio-temporal information in a high-dimensional feature space to form complementary "where-how" relationships at the semantic level, thereby providing a richer semantic information for the model. Extensive experimentation on the CASME \uppercase\expandafter{\romannumeral2}, SAMM and CAS(ME)$^3$ datasets has demonstrated that TSFmicro exhibits superior performance in micro-expression recognition tasks when compared to existing state-of-the-art methods.

Notwithstanding the considerable progress accomplished by TSFmicro, a plethora of unresolved issues persist within the domain of dynamic microexpression recognition. A salient issue pertains to the intricacies inherent in cross-cultural scenarios. For instance, the SAMM dataset encompasses a broad spectrum of cultural backgrounds. In this context, while the performance of TSFmicro is optimal in comparison to other state-of-the-art methods, there is considerable scope for enhancement when evaluated against the performance results in the CASME \uppercase\expandafter{\romannumeral2} dataset. This suggests that the current method has a generalization bottleneck in coping with different cultural expression habits and facial morphology differences, and the model may have difficulty in adapting to changes in micro-expression features due to cultural differences in cross-cultural scenarios. Another major issue is the imbalance of categories. In the micro-expression dataset, the number of negative emotion samples is low, which leads to a lower accuracy of the model in recognizing negative emotions. This problem is especially pronounced in the SAMM dataset and may further restrict the model's performance in practical applications. In order to address these issues, future research could explore methods such as transfer learning and contrast learning to improve the model's ability to generalise to cross-cultural scenarios and handle imbalanced data.

Furthermore, future research could explore the application of TSFmicro to domains other than micro-expression recognition, such as psychological state monitoring or expression analysis in the security domain. Concomitantly, further research into the applicability of the framework in diverse cultural contexts and the interpretability of the learned features will facilitate the acquisition of further insights and improvements.

\section*{Acknowledgment}

This research is supported by the National Key R\&D Program "Active Health and Aging Population Technology Response" Key Project(No.2024YFC3606802), and also supported by the Beijing Key Laboratory of Behavior and Mental Health, Peking University.

% \section*{Declaration of generative AI and AI-assisted technologies in the writing process}

% During the preparation of this work, the author(s) used ChatGPT to enhance the clarity of the language. After using this tool, the author(s) reviewed and edited the content as needed and take(s) full responsibility for the content of the publication.

% \newpage

\printcredits

\bibliographystyle{unsrt}

\bibliography{cas-refs}

\end{document}